\documentclass{article}

\PassOptionsToPackage{numbers,sort&compress}{natbib} 
\usepackage[final]{neurips_2024_mod}
\usepackage{amsmath}
\usepackage[utf8]{inputenc} 
\usepackage[T1]{fontenc}    
\usepackage{hyperref}       
\usepackage{url}            
\usepackage{booktabs}       
\usepackage{amsfonts}       
\usepackage{nicefrac}       
\usepackage{microtype}      
\usepackage{xcolor}         
\usepackage{graphicx}
\usepackage{algorithm}
\usepackage{algpseudocode}
\usepackage{xfrac}
\usepackage[all]{hypcap}

\title{Neural Attention: A Novel Mechanism for Enhanced Expressive Power in Transformer Models}

\author{%
  Andrew J.~DiGiugno\\
  Department of Computer Science\\
  University of Bridgeport\\
  Bridgeport, CT 06604 \\
  \texttt{adigiugn@my.bridgeport.edu} \\
  \And
  Ausif Mahmood\\
  Department of Computer Science\\
  University of Bridgeport\\
  Bridgeport, CT 06604 \\
  \texttt{mahmood@bridgeport.edu} \\
}

\begin{document}

\maketitle

\renewcommand{\thefootnote}{}
\footnotetext{Preprint. This work has not been peer-reviewed.}
\renewcommand{\thefootnote}{\arabic{footnote}}

\begin{abstract}
  Transformer models typically calculate attention matrices using dot products, which have limitations when capturing nonlinear relationships between embedding vectors. We propose \textit{Neural Attention}, a technique that replaces dot products with feed-forward networks, enabling a more expressive representation of relationships between tokens. This approach modifies only the attention matrix calculation while preserving the matrix dimensions, making it easily adaptable to existing transformer-based architectures. We provide a detailed mathematical justification for why Neural Attention increases representational capacity and conduct controlled experiments to validate this claim. When comparing Neural Attention and Dot-Product Attention, NLP experiments on WikiText-103 show a reduction in perplexity of over 2 percent. Similarly, experiments on CIFAR-10 and CIFAR-100 show improvements in accuracy of more than 4 percentage points for image classification tasks. While Neural Attention introduces higher computational demands, we develop techniques to mitigate these challenges, ensuring practical usability without sacrificing the increased expressivity it provides. This work establishes Neural Attention as an effective means of enhancing the predictive capabilities of transformer models across a variety of applications. The code for all experiments is available at \url{https://github.com/awayfromzel/neural-attention-research}.
\end{abstract}

\section{Introduction}
\label{sec:introduction}
Transformers have revolutionized artificial intelligence, delivering groundbreaking advances in Natural Language Processing (NLP) and computer vision tasks. At the core of these models lies the attention mechanism, which captures relationships between embedding vectors that represent data with spatial dependencies. The most widely used implementation, Dot-Product Attention, is foundational to the success of transformers~\cite{vaswani2017attention}. However, it operates within a constrained representational space, limiting its ability to model nonlinear relationships and reducing the expressivity of the model.

We introduce \textit{Neural Attention}, a novel technique designed to enhance the expressive capacity of transformers. By replacing dot products with learnable weight matrices in feed-forward networks, Neural Attention enables transformers to better capture intricate, nonlinear dependencies between embedding vectors. Unlike most current research, which is focused on improving the computational efficiency of attention mechanisms, our method prioritizes enhancing their representational power, which is a critical factor in advancing the foundational capabilities of transformers. In this work, we clearly demonstrate this added expressivity on both NLP and vision tasks. While Neural Attention exhibits higher computational and memory requirements, we develop an implementation that significantly reduces this additional overhead, as detailed in Section~\ref{sec:methodology}.

\subsection{Transformer Basics}
\label{subsec:transformer_basics}
A key challenge in modeling sequential data is that the meaning of each data point is dependent on the data that surrounds it. For example, in text, the meaning of each word depends on the context provided by surrounding words. In an image, the meaning of each pixel depends on the surrounding pixels. Transformers address this challenge by encoding data sequences of length \( n \), into embedding vectors of length \( d \). These embedding vectors become contextually aware representations of each element in the sequence and are arranged into three learnable matrices: Query (\( \mathit{Q} \)), Key (\( \mathit{K} \)), and Value (\( \mathit{V} \)), each with size \( \mathbb{R}^{n \times d} \). Each matrix can be viewed as a different representation of the same input sequence.  An "attention" operation is performed between $\mathit{Q}$ and $\mathit{K}$, which creates the attention matrix $\mathit{A} \in \mathbb{R}^{n \times n}$. This is a matrix of "attention scores", which quantifies the importance of each token in the sequence with respect to the others~\cite{vaswani2017attention}. This attention matrix is then used to weight the \( \mathit{V} \) matrix, producing a new representation of the sequence as: $\textit{softmax}\left(\smash{\sfrac{\mathit{A}}{\sqrt{d}}}\right)\!\mathit{V} \in \mathbb{R}^{n \times d}$. This process is repeated through multiple layers, with each layer refining the embeddings to capture increasingly complex relationships. An output layer is then used to make a prediction. Traditionally, the attention scores in matrix $\mathit{A}$ are computed using the matrix product between $\mathit{Q}$ and $\mathit{K^\top}$, as shown in Equation~\ref{eq:scaled_dot_product_attention}. The key contribution of this work is to explore a different method for calculating this matrix, which allows for a better contextual understanding of the data.

\begin{align}
    \text{Scaled Dot-Product Attention}(Q, K, V) &= \text{Softmax} \left( \frac{Q K^\top}{\sqrt{d}} \right) V \label{eq:scaled_dot_product_attention}
\end{align}

\subsection{Theoretical Justification of Neural Attention}
\label{subsec:theoretical_justification}
In Equation~\ref{eq:dot_product}, we demonstrate a vector $\vec{q} \in \mathbb{R}^{d \times 1}$ from the \textbf{\textit{$Q$}} matrix, and a vector $\vec{k} \in \mathbb{R}^{d \times 1}$ from the \textbf{\textit{$K$}} matrix, producing an attention score through a dot product operation. Note that it is only possible for elements of the same index to interact when mapping two vectors to a scalar in this way. 
\begin{align}
    \text{Attention Score} &= \sum_{i=1}^{d} \vec{q}_i \cdot \vec{k}_i \label{eq:dot_product}
\end{align}
By using the dot product, we are creating an attention score that measures global, linear dependencies between embedding vectors. This potentially misses relationships between individual elements, especially those at different indices. With Dot-Product Attention, transformers must learn embeddings that are capable of expressing all inter-dependencies through a dot product. In contrast, Neural Attention allows the model to learn a unique function that maps the two embedding vectors to a scalar. This makes it capable of modeling nonlinear dependencies that a dot product cannot capture. It does this by first concatenating them into the vector $\vec{qk}_{\text{concat}} \in \mathbb{R}^{2d \times 1}$. We then pass $\vec{qk}_{\text{concat}}$ into a hidden layer using a learnable weight matrix $\mathit{W_h} \in \mathbb{R}^{h \times 2d}$ and bias $\vec{b}_h \in \mathbb{R}^{h \times 1}$, where $h$ is the number of neurons in the hidden layer. The vector $\vec{h} \in \mathbb{R}^{h \times 1}$  encodes information about dependencies between $\vec{q}$ and $\vec{k}$. This can be seen in Equation~\ref{eq:hidden_vector}, where $\sigma$ represents a nonlinear activation function.
\begin{align}
    \vec{h} &= \sigma\left(\mathit{W_h} \cdot \text{concat}(\vec{q}, \vec{k}) + \vec{b}_h\right) \label{eq:hidden_vector}
\end{align}

The hidden vector $\vec{h}$ is then projected to a scalar using a learnable vector of weights $\vec{w}_a^\top \in \mathbb{R}^{1 \times h}$ and bias $b_a$. This produces an attention score that allows the model to account for local, nonlinear dependencies between embedding vectors, as shown in Equation~\ref{eq:neural_attention_score}. A visual representation comparing the calculation of Dot-Product Attention and Neural Attention can be seen in Appendix~\ref{sec:Illustrations}.
\begin{align}
    \text{Attention Score} &= \vec{w}_a^\top \sigma\left(\mathit{W_h} \cdot \text{concat}(\vec{q}, \vec{k}) + \vec{b}_h\right) + b_a \label{eq:neural_attention_score}
\end{align}

We illustrate the geometric differences between Dot-Product Attention and Neural Attention by considering an example with an embedding dimension of 2, where $\vec{q} = [q_1, q_2]$ and $\vec{k} = [k_1, k_2]$. In this case, \((\vec{q}, \vec{k}) \in \mathbb{R}^2 \times \mathbb{R}^2\), and the scalar output \(z \in \mathbb{R}\). For $f(\vec{q},\vec{k})=z$, this defines the mapping:
\[
f: \mathbb{R}^2 \times \mathbb{R}^2 \to \mathbb{R}
\]

\begin{figure}[H]
    \centering
    \begin{minipage}{0.45\textwidth} 
        \centering
        \textbf{\(z = x^2 + y^2\)} 
        \\[0.5em] 
        \vspace{1em} 
        \hspace*{-1.5cm} 
        \includegraphics[scale=0.43]{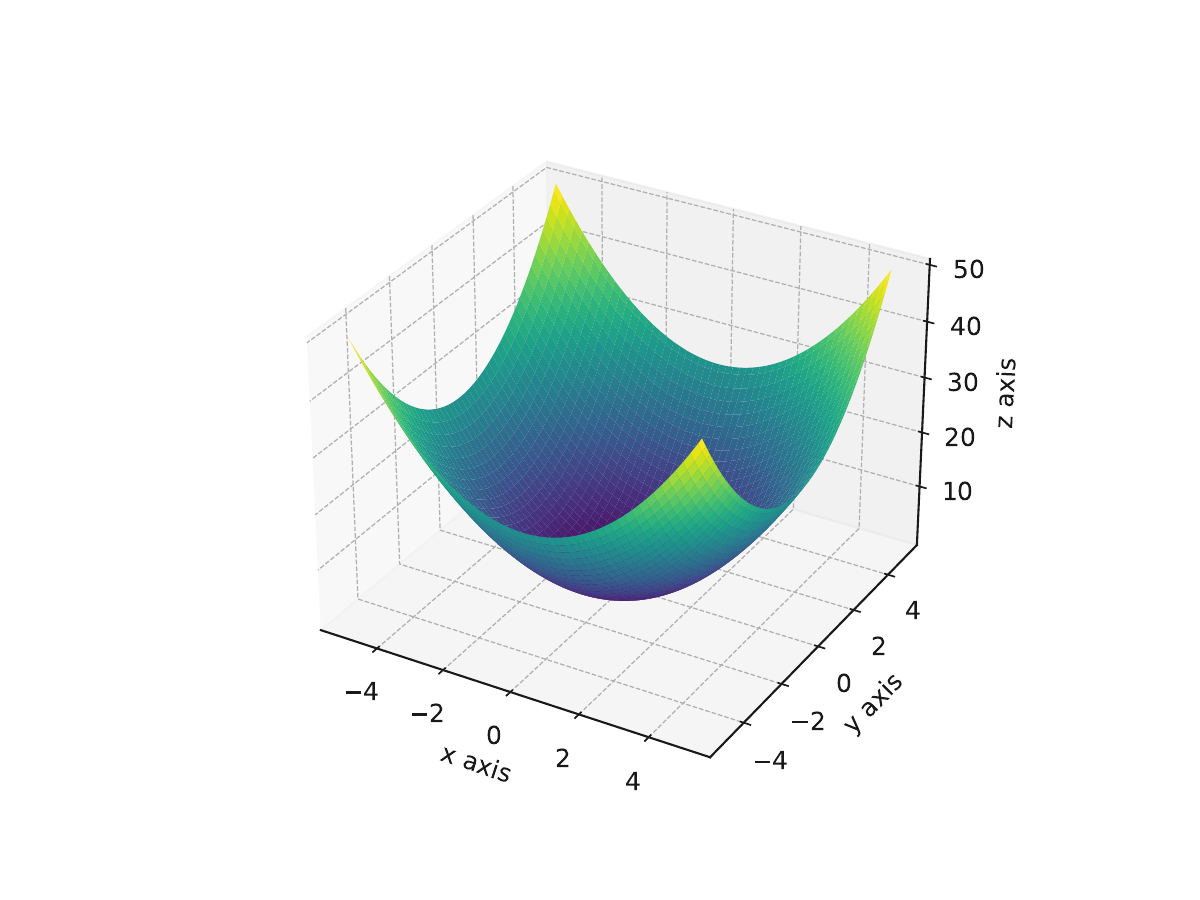} 
    \end{minipage}
    \hfill 
    \begin{minipage}{0.45\textwidth} 
        \centering
        \textbf{\(z = \vec{w}_a^\top \sigma\left(\mathit{W_h} \cdot 
        \begin{bmatrix}
        x \\ y \\ x \\ y
        \end{bmatrix}
        + \vec{b}_h
         \right) + b_a\)} 
        \\[0.5em] 
        \hspace*{-1.5cm} 
        \includegraphics[scale=0.43]{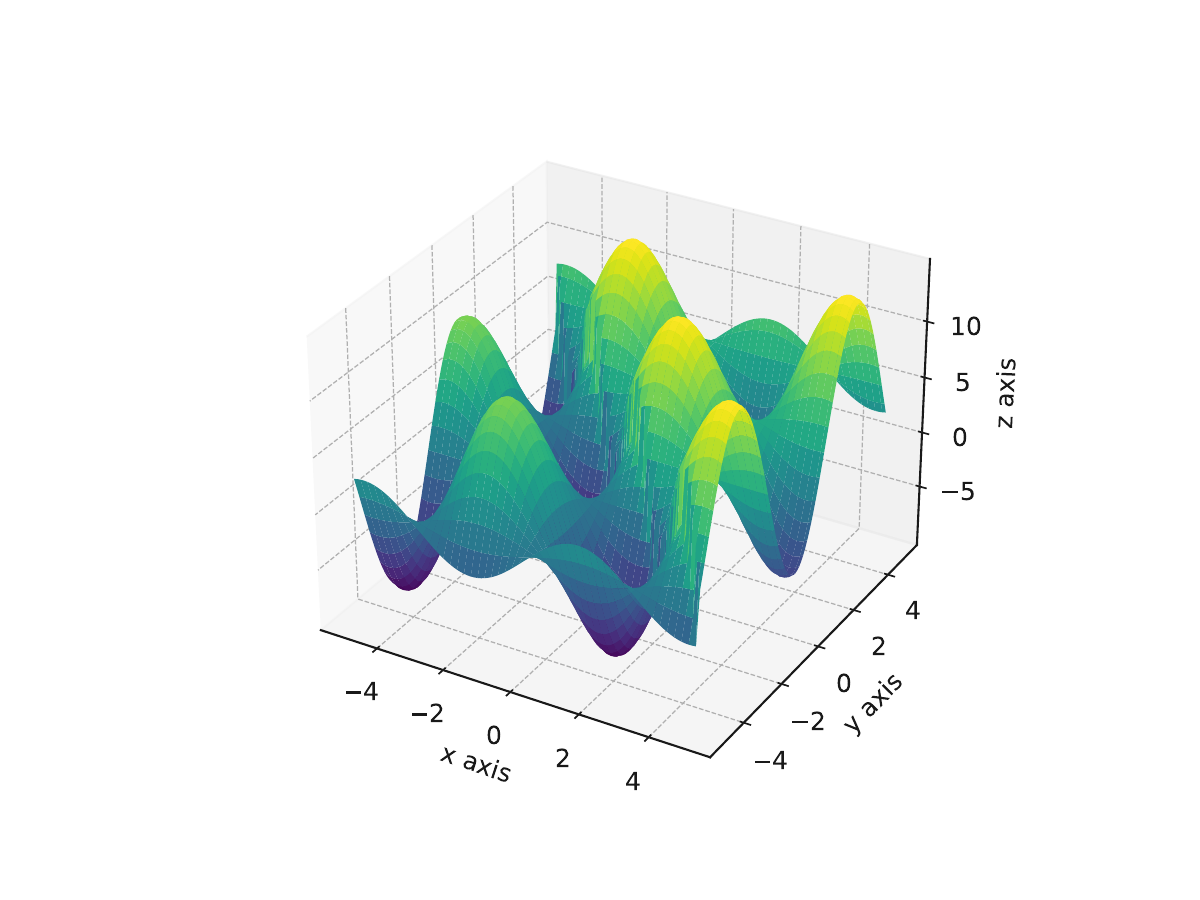} 
    \end{minipage}
    \caption{A smooth parabolic surface (left) versus a complex surface with sharp edges (right).}
    \label{fig:three_d_manifolds}
\end{figure}

When including both the input embeddings and the scalar output, we consider the 5-dimensional space \(\mathbb{R}^5\), where each point is represented as \((q_1, q_2, k_1, k_2, z)\). By restricting the embeddings to the case \(q_1 = k_1 = x\) and \(q_2 = k_2 = y\), we can visualize the attention as a 3D manifold embedded in \(\mathbb{R}^5\), as seen in Figure~\ref{fig:three_d_manifolds}. In this case, Dot-Product Attention is given by $z = \vec{q} \cdot \vec{k} = x^2 + y^2$, which corresponds to a smooth, continuous 3D manifold, embedded in \(\mathbb{R}^5\). Specifically, it is a paraboloid. In contrast, Neural Attention introduces a nonlinear mapping:
\[
z = \vec{w}_a^\top \sigma\left(\mathit{W_h} \cdot 
\text{concat}(\vec{q}, \vec{k})
+ \vec{b}_h
\right)
+ b_a
=
\vec{w}_a^\top \sigma\left(\mathit{W_h} \cdot 
\begin{bmatrix}
x \\
y \\
x \\
y
\end{bmatrix}
+ \vec{b}_h
\right)
+ b_a
\]
This enables \(z\) to exist on a far more flexible 3D manifold in \(\mathbb{R}^5\), defined by the learned parameters \(\mathit{W_h}\), $\vec{w}_a^\top$, $\vec{b}_h$, and $b_a$. Unlike the paraboloid produced by Dot-Product Attention, this manifold can capture intricate, nonlinear relationships, including sharp transitions and complex geometries. In Figure~\ref{fig:three_d_manifolds}, the left plot shows the smooth paraboloid surface modeled by Dot-Product Attention and the right plot illustrates the potential complexity of a manifold modeled by Neural Attention.

\section{Related Work}
\label{sec:related_work}
The attention mechanism was first introduced within the context of machine translation by~\citet{bahdanau2014neural}, when it was used for dynamic alignment of words between input and output sequences. This work introduced the concept of "additive attention", in which two vectors are independently processed by their own weight matrices to produce vectors that are summed and used to calculate the final attention score. With Neural Attention, instead of processing each vector separately, we concatenate the two vectors and perform a nonlinear operation on the combined representation, projecting them jointly into a space that encodes their relationship. This approach captures richer dependencies between vectors, whereas additive attention ultimately relies on a linear combination of independently projected representations.

Building upon this foundation,~\citet{vaswani2017attention} proposed the transformer architecture, eliminating the need for recurrence while still allowing flexibility of input sequence length. This innovation significantly improved the ability of sequence processing to be parallelized, addressing some of the bottlenecks in recurrent models. In this work, they adopted Dot-Product Attention as the mechanism for calculating attention scores, citing its practical advantages over additive attention when considering computational and memory efficiency. However, their work did not explore whether additive attention could offer advantages in expressivity, leaving the representational limitations of Dot-Product Attention largely unexamined. Neural Attention addresses this gap by prioritizing enhanced expressivity, even as we tackle the associated complexity challenges, as explained in Section~\ref{sec:methodology}.

While these foundational innovations laid the groundwork for modern transformers, much of the subsequent research has shifted toward improving computational efficiency, rather than addressing representational limitations. FastFormer~\cite{wu2021fastformer} explored additive attention to achieve faster processing, while SwiftFormer~\cite{shaker2023swiftformer} revisited additive attention for real-time applications in mobile vision tasks. Similarly,~\citet{zhang2024cas} proposed additive and convolution-based attention mechanisms tailored for vision transformers, focusing on practical implementation benefits. Linformer~\cite{wang2020linformer} addressed the $O(n^2)$ complexity of self-attention by employing low-rank approximations to reduce computational and memory costs. More recently,~\citet{mahmood2024enhanced} introduced a segmented attention approach, computing attention only on pairs of consecutive overlapping segments to further optimize computational efficiency. While these efforts underscore the ongoing interest in rethinking attention mechanisms, they largely prioritize efficiency improvements over representational power. Neural Attention distinguishes itself by directly addressing the expressivity limitations of standard attention mechanisms, providing a novel approach to modeling relationships between embedding vectors. 

In the current literature, there are not many examples of work solely focused on improving the expressive power of transformers by exploring alternative approaches to the attention mechanism. However, RoFormer \cite{su2024roformer} improved the capability of transformers to capture long-term dependencies by introducing a novel technique for preserving relative position relationships in self-attention calculations. gMLP \cite{liu2021pay} replaced the attention mechanism entirely by using multi-layer perceptrons with gating. This work relates to ours in that it uses feed-forward networks to learn spatial dependencies. This provides a justification for our work; however, it is a fundamentally different approach and cannot be easily adapted into existing attention-based transformer architectures, as ours can be. Moreover, their implementation does not allow for nor was it tested on autoregressive tasks. The work most similar to ours is M{\"o}bius Attention \cite{halacheva2024expanding} which sought to improve the expressive power of attention mechanisms by adding a nonlinear M{\"o}bius transformation and calculating attention scores in a complex space. This differs from our work in that they only apply the nonlinear transformation to the embedding vectors in the $\mathit{Q}$ matrix, creating a richer representation prior to attention calculation. Neural Attention, by contrast, adds a nonlinear transformation into the attention calculation itself.

\section{Methodology}
\label{sec:methodology}

\subsection{Implementation of Neural Attention}
\label{subsec:implementation}
We begin with the Query ($\mathit{Q}$) and Key ($\mathit{K}$) matrices, commonly used in transformer models. We define these matrices to have a sequence length $n$ and an embedding dimension $d$.
A linear down-projection is performed on both matrices along the embedding dimension to make subsequent steps less resource-intensive. We define the resulting matrices as $\mathit{Q}^{\prime}$ and $\mathit{K}^{\prime}$, and the reduced embedding dimension as $d^{\prime}$. This dimensionality reduction is shown below in Equations~\ref{eq:Qprime} and~\ref{eq:Kprime}.
\begin{align}
    \mathit{Q}^{\prime} &= \mathit{Q} \mathit{W}_q, \quad \mathit{Q} \in \mathbb{R}^{n \times d}, \quad \mathit{W}_q \in \mathbb{R}^{d \times d'}, \quad \mathit{Q}^{\prime} \in \mathbb{R}^{n \times d'} \label{eq:Qprime}\\
    \mathit{K}^{\prime} &= \mathit{K} \mathit{W}_k, \quad \mathit{K} \in \mathbb{R}^{n \times d}, \quad \mathit{W}_k \in \mathbb{R}^{d \times d'}, \quad \mathit{K}^{\prime} \in \mathbb{R}^{n \times d'} \label{eq:Kprime}
\end{align}
We then reshape $\mathit{Q}^{\prime}$, making it a tensor by adding a singleton dimension and giving it the form
$\mathbf{Q^{\prime}} \in \mathbb{R}^{n \times 1 \times d'}$. We reshape $\mathit{K^{\prime}}$, making it a tensor by adding a singleton dimension and giving it the form $\mathbf{K^{\prime}} \in \mathbb{R}^{1 \times n \times d'}$. By broadcasting along the singleton dimensions, we get the resulting forms: $\mathbf{Q^{\prime}} \in \mathbb{R}^{n \times n \times d'}$ and $\mathbf{K^{\prime}} \in \mathbb{R}^{n \times n \times d'}$. Below, this process is written step by step.
\begin{align*}
\mathit{Q}^{\prime} &\in \mathbb{R}^{n \times d'} \quad \xrightarrow{\text{reshape}} \quad \mathbf{Q}^{\prime} \in \mathbb{R}^{n \times 1 \times d'} \quad \xrightarrow{\text{broadcast}} \quad \mathbf{Q}^{\prime} \in \mathbb{R}^{n \times n \times d'} \\
\mathit{K}^{\prime} &\in \mathbb{R}^{n \times d'} \quad \xrightarrow{\text{reshape}} \quad \mathbf{K}^{\prime} \in \mathbb{R}^{1 \times n \times d'} \quad \xrightarrow{\text{broadcast}} \quad \mathbf{K}^{\prime} \in \mathbb{R}^{n \times n \times d'}
\end{align*}

\begin{figure}[H]
    \centering
    \includegraphics[width=\textwidth]{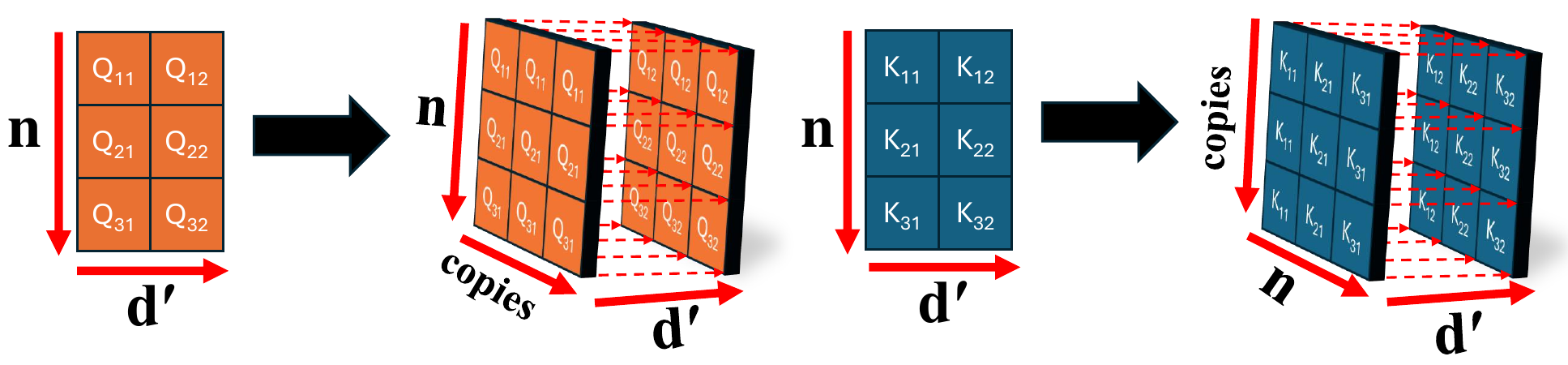}
    \caption{Reshape and broadcast process of  matrices $\mathit{Q^{\prime}}$ and $\mathit{K^{\prime}}$ into tensors $\mathbf{Q^{\prime}}$ and $\mathbf{K^{\prime}}$. Prime symbols are left out for better readability.}
\label{fig:reshape_and_broadcast}
\end{figure}

\begin{figure}[H]
    \centering
      \includegraphics[scale=0.75]{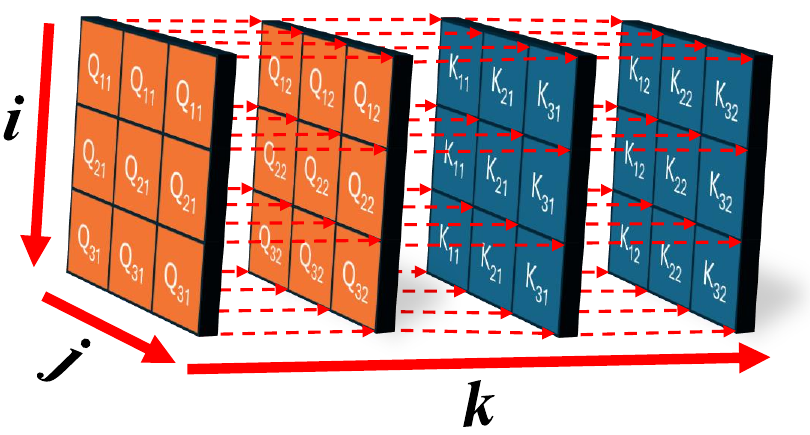}
    \caption{Tensor $\mathbf{C}$, created after concatenating tensors $\mathbf{Q^{\prime}}$ and $\mathbf{K^{\prime}}$ from Figure~\ref{fig:reshape_and_broadcast} along their embedding dimension.}
    \label{fig:c_tensor}
\end{figure}

In Figure~\ref{fig:reshape_and_broadcast}, we demonstrate matrices $\mathit{Q^{\prime}}$  (shown in orange) and $\mathit{K^{\prime}}$ (shown in blue) undergoing the reshape and broadcast process and becoming tensors. These matrices are defined to have $n=3$ and $d^{\prime}=2$. The red arrows signify what data is contained in each dimension. Indices are included for clarity. Note that broadcasting along perpendicular axes can be seen as analogous to transposing the $\mathit{K'}$ matrix in Dot-Product Attention implementations. This is represented by the "copies" axis.
After the reshape and broadcast steps, we then create a tensor $\mathbf{C}$ by concatenating the $\mathbf{Q^{\prime}}$ and $\mathbf{K^{\prime}}$ tensors along their embedding dimension. The tensor resulting from applying Equation~\ref{eq:C_Matrix} to the $\mathbf{Q^{\prime}}$ and $\mathbf{K^{\prime}}$ tensors in Figure~\ref{fig:reshape_and_broadcast} is shown in Figure~\ref{fig:c_tensor}. The axes $i$, $j$, and $k$ are added for clarity, as they will be used in subsequent equations to index this tensor.
\begin{align}
    \mathbf{C} &= \text{concat}(\mathbf{Q'}, \mathbf{K'}, \text{axis}=-1), \quad \mathbf{C} \in \mathbb{R}^{n \times n \times 2d'} \label{eq:C_Matrix}
\end{align}
We then utilize the same learnable parameters discussed in Section~\ref{subsec:theoretical_justification}, $\mathit{W}_h \in \mathbb{R}^{h \times 2d'}$, $\vec{b}_h \in \mathbb{R}^{h \times 1}$, $\vec{w}_a^\top \in \mathbb{R}^{1 \times h}$ and $b_a$ to calculate the final attention matrix. Every $(i,j)$-th slice of tensor $\mathbf{C}$ along the $\mathit{k}$ axis, $\mathbf{C}_{ij}$, shares these same parameters when calculating its attention score. The calculation of a single attention score is given below by Equation~\ref{eq:Attention_Matrix_Individual}. 
 \begin{align}
\text{AttentionScore}_{ij} = \vec{w}_a^\top \sigma(\mathit{W}_h \mathbf{C}_{ij} + \vec{b}_h) + b_a \label{eq:Attention_Matrix_Individual}
\end{align}

More generally, after processing tensor $\mathbf{C}$ using Equation~\ref{eq:Attention_Matrix}, we define the resulting matrix of attention scores as $\mathit{A}$. There is a visual example using this equation given in Appendix~\ref{sec:Example}.
 \begin{align}
\mathit{A} = \vec{w}_a^\top \sigma(\mathit{W}_h \mathbf{C} + \vec{b}_h) + b_a, \quad \mathit{A} \in \mathbb{R}^{n \times n} \label{eq:Attention_Matrix}
\end{align}

The final calculation for Neural Attention is given by Equation~\ref{eq:neural_attention_final}. Note that the only difference between this and the Scaled Dot-Product Attention, seen in Equation~\ref{eq:scaled_dot_product_attention}, is the numerator in the softmax function. Also note that $\mathit{A}$ can be derived from $\mathit{Q}$ and $\mathit{K}$. Both of these equations produce a matrix of size $n \times n$, making Neural Attention easy to implement in any architecture currently using Dot-Product Attention.

\begin{align}
    \text{Neural Attention}(Q, K, V) &= \text{Softmax} \left( \frac{A}{\sqrt{d}} \right) V \label{eq:neural_attention_final}
\end{align}

To provide further clarity, the below pseudocode shows how the attention matrix is calculated in our implementation of Neural Attention.

\begin{algorithm}[H]
\caption{Calculating Attention Scores}
\label{alg:neural_attention}
\begin{algorithmic}[1]
\Require Query matrix $Q \in \mathbb{R}^{\text{batch} \times \text{heads} \times \text{seq\_length} \times \text{embedding\_dim}}$
\Require Key matrix $K \in \mathbb{R}^{\text{batch} \times \text{heads} \times \text{seq\_length} \times 
\text{embedding\_dim}}$
\Ensure Attention scores $A \in \mathbb{R}^{\text{batch} \times \text{heads} \times \text{seq\_length} \times \text{seq\_length}}$
\State Perform linear down-projection of $Q$ and $K$ along the embedding dimension:
    \[
    Q' = Q W_q, \quad W_q \in \mathbb{R}^{\text{embedding\_dim} \times \text{reduced\_dim}}, \quad Q' \in \mathbb{R}^{\text{batch} \times \text{heads} \times \text{seq\_length} \times \text{reduced\_dim}}
    \]
    \[
    K' = K W_k, \quad W_k \in \mathbb{R}^{\text{embedding\_dim} \times \text{reduced\_dim}}, \quad K' \in \mathbb{R}^{\text{batch} \times \text{heads} \times \text{seq\_length} \times \text{reduced\_dim}}
    \]
\State Reshape $Q'$ and $K'$: Add singleton dimensions and repeat along sequence length
    \[
        Q' \xrightarrow{\text{reshape}} \mathbb{R}^{\text{batch} \times \text{heads} \times \text{seq\_length} \times 1 \times \text{reduced\_dim}}
        \xrightarrow{\text{broadcast}} \mathbb{R}^{\text{batch} \times \text{heads} \times \text{seq\_length} \times \text{seq\_length} \times \text{reduced\_dim}}
    \]
    \[
        K' \xrightarrow{\text{reshape}} \mathbb{R}^{\text{batch} \times \text{heads} \times 1 \times \text{seq\_length} \times \text{reduced\_dim}}
        \xrightarrow{\text{broadcast}} \mathbb{R}^{\text{batch} \times \text{heads} \times \text{seq\_length} \times \text{seq\_length} \times \text{reduced\_dim}}
    \]
\State Concatenate $\mathbf{Q'}$ and $\mathbf{K'}$ along the embedding dimension (axis=-1):
    \[
    \text{concat}(\mathbf{Q'}, \mathbf{K'}, \text{axis}=-1) \rightarrow \mathbf{C} \in \mathbb{R}^{\text{batch} \times \text{heads} \times \text{seq\_length} \times \text{seq\_length} \times 2 \cdot \text{reduced\_dim}}
    \]
\State Pass all vectors in the final dimension of $\mathbf{C}$ through their own feed-forward networks:
   \Statex \hspace{1em} Hidden Layer: Apply a linear transformation and nonlinear activation function
    \Statex \hspace{1em} Output Layer: Apply a linear transformation to reduce the final dimension to 1
\State Squeeze the final dimension to output attention scores $A$: 
    \[
        A \in \mathbb{R}^{\text{batch} \times \text{heads} \times \text{seq\_length} \times \text{seq\_length} \times \text{1}}\xrightarrow{\text{squeeze}} A \in \mathbb{R}^{\text{batch} \times \text{heads} \times \text{seq\_length} \times \text{seq\_length}}
    \]
\end{algorithmic}
\end{algorithm}

\subsection{Overcoming Complexity Challenges}
\label{subsec:overcocming_complexity_challenges}
Both Dot-Product Attention and Neural Attention have identical time complexities of $O(n^{2})$ with respect to the sequence length. However, Neural Attention has a significantly higher memory requirement due to the large intermediate tensor $\mathbf{C}$, seen in Equation~\ref{eq:C_Matrix}. To overcome this challenge, Neural Attention is used only in the first layer of our transformer model, with all subsequent layers using Dot-Product Attention. This approach is used in combination with down-projection, as seen in Equations~\ref{eq:Qprime} and~\ref{eq:Kprime}. In our testing, we find that even when projecting to an embedding dimension as low as $d^{\prime}=2$, Neural Attention produces better results than when no down-projection is used at all. This implies that Neural Attention is low-rank, able to make meaningful contributions to the model with a small embedding dimension. For these reasons, it is possible to take advantage of the increased expressive power of this technique in a scalable manner, as evidenced in Section~\ref{subsec:computational_and_memory_comparison}.

\section{Experiments and Results}
\label{sec:experiments_and_results}

The effectiveness of Neural Attention is tested in both a generative NLP context and an image classification context. This is beneficial because it provides a diversity of data types and training modalities. The former requires attention to be used in an autoregressive context, while the latter does not. In every test presented in this section, all training, architecture, and data are identical for both Neural Attention and Dot-Product Attention. The only difference being that Neural Attention is applied in the first layer only, with subsequent layers using Dot-Product Attention, as detailed in Section~\ref{subsec:overcocming_complexity_challenges}. None of the models use pretraining, all are trained from scratch. In these results, the focus is not on raw performance indicators (perplexity for NLP and accuracy for image classification), but rather on the \textit{relative} improvement seen when using Neural Attention vs Dot-Product Attention. All tests were performed using an NVIDIA RTX 4090 with 24GB of VRAM. 

\subsection{Generative NLP Testing}
\label{subsec:NLP_Testing}
The tests in this section are performed using a decoder-only, multi-head, causal self-attention transformer model, similar to that described by Radford et al.~\cite{radford2018improving}. Training is performed in an autoregressive manner and utilizes sinusoidal positional encoding. All tests use 8 transformer layers, each one having 8 attention heads. A sequence length of 1024 and a batch size of 16 is used. Generative performance is benchmarked using perplexity measurements every 10,000 iterations while training on the WikiText-103 dataset~\cite{merity2016pointer} for 1,000,000 iterations in total. An embedding dimension of 512 is used for tokens and an embedding dimension of $d=64$ per head is used in the $Q$, $K$, and $V$ matrices. Neural Attention is tested without the down-projection described by Equations~\ref{eq:Qprime} and~\ref{eq:Kprime}, as well as with a value of $d^{\prime}=16$ and a value $d^{\prime}=2$. Note that when no down-projection is used, we do not apply Equations~\ref{eq:Qprime} and~\ref{eq:Kprime} with a value of $d^{\prime}=d$, but instead skip this step entirely, substituting out $Q^{\prime}$ and $K^{\prime}$ for $Q$ and $K$. Training graphs for all tests can be found in Appendix~\ref{sec:graphs}.

\begin{table}[h!]
    \centering
    \caption{Comparison of perplexity results after 1 million iterations.}
    \label{tab:perplexity_results}
    \resizebox{\textwidth}{!}{
        \begin{tabular}{|c|c|c|c|}
            \hline
            \textbf{Method} & \textbf{Model Size (params)} & \textbf{Down-Projection ($d'$)} & \textbf{Lowest Perplexity} \\ \hline
            Dot-Product Attention & 46.528M & None ($d=64$) & 26.03 \\ \hline
            Neural Attention  & 46.528M & 2 & 25.46 \\ \hline
            Neural Attention  & 46.531M & 16 & 25.64 \\ \hline
            Neural Attention & 46.538M  & None ($d=64$)  & 25.56 \\ \hline        
        \end{tabular}
    }
    \label{tab:perplexity_results}
\end{table}
\begin{table}[h!]
    \centering
    \caption{Percent improvement in perplexity over Dot-Product Attention for various Neural Attention configurations. Derived from the data in Table~\ref{tab:perplexity_results}.}
    \label{tab:percent_improvement}
    \begin{tabular}{|c|c|}
        \hline
        \textbf{Down-Projection ($d'$)} & \textbf{Percent Improvement Over Dot-Product Attention (\%)} \\ \hline
        $2$                          & 2.19\% \\ \hline
        $16$                         & 1.49\% \\ \hline
        None ($d=64$)                & 1.81\% \\ \hline
    \end{tabular}
\end{table}

Table~\ref{tab:percent_improvement} shows that the use of Neural Attention leads to a significant improvement in perplexity, with the best result being a reduction of $2.19\%$. One should pay particular attention to the fact that down-projecting the $\mathit{Q}$ and $\mathit{K}$ matrices does not negatively impact performance. Interestingly, the best-performing test case was the one that used the highest amount of dimensionality reduction. This could potentially be explained by the additional linear layer and learnable parameters introduced by the projection or by a beneficial dropout effect.

\subsection{Image Classification Testing}
\label{subsec:image_classification_results}

These tests are performed using a vision transformer similar to that described by Dosovitskiy et al. \cite{dosovitskiy2020image}, with a patch size of 8 pixels, an embedding dimension of 768, and a batch size of 32. The images are resized to 224 by 224, giving a sequence length of $28^2 + 1 = 785$. The model has 12 layers, each with 8 attention heads, utilizing self-attention with sinusoidal positional encoding and no masking. The accuracy of image classification is benchmarked using the CIFAR-10 and CIFAR-100 datasets~\cite{krizhevsky2009learning}. Note that CIFAR-10 creates a 10-class classification problem, while CIFAR-100 creates a 100-class classification problem.

\begin{table}[h!]
    \centering
    \caption{Validation accuracy after 350 training epochs on CIFAR-10.}
    \label{tab:accuracy_results}
    \resizebox{\textwidth}{!}{
        \begin{tabular}{|c|c|c|c|}
            \hline
            \textbf{Method} & \textbf{Model Size (params)} & \textbf{Down-Projection ($d'$)} & \textbf{Validation Accuracy (\%)} \\ \hline
            Dot-Product Attention & 47.402M & None ($d=64$) & 83.70 \\ \hline
            Neural Attention & 47.472M & 2 & 86.70 \\ \hline
        \end{tabular}
    }
\end{table}

\begin{table}[h!]
    \centering
    \caption{Validation accuracy after 350 training epochs on CIFAR-100.}
    \label{tab:accuracy_results_cifar100}
    \resizebox{\textwidth}{!}{
        \begin{tabular}{|c|c|c|c|}
            \hline
            \textbf{Method} & \textbf{Model Size (params)} & \textbf{Down-Projection ($d'$)} & \textbf{Validation Accuracy (\%)} \\ \hline
            Dot-Product Attention & 47.402M & None ($d=64$) & 55.63 \\ \hline
            Neural Attention & 47.472M & 2 & 59.89 \\ \hline
        \end{tabular}
    }
\end{table}

\begin{table}[H]
    \centering
    \caption{Improvement in validation accuracy with Neural Attention ($d'=2$) compared to Dot-Product Attention.}
    \label{tab:accuracy_improvement_simple}
    \begin{tabular}{|c|c|}
        \hline
        \textbf{Dataset} & \textbf{Percentage Point Improvement over Dot-Product Attention} \\ \hline
        CIFAR-10  & 3.0 \% \\ \hline
        CIFAR-100 & 4.26 \% \\ \hline
    \end{tabular}
\end{table}

\label{others}

\subsection{Computational and Memory Comparison}
\label{subsec:computational_and_memory_comparison}
Memory requirements and inference speeds during training for all experiments are in the below tables. These metrics are included to provide a comprehensive picture of Neural Attention's implementation, but the primary takeaway remains its superior expressivity, as evidenced by the results in Sections~\ref{subsec:NLP_Testing} and~\ref{subsec:image_classification_results}. Techniques to address the complexity challenges associated with Neural Attention, discussed in section~\ref{subsec:overcocming_complexity_challenges}, have achieved a reasonable memory requirement and inference speed when using a value of $d'=2$, making Neural Attention scalable. Note that with Neural Attention only being applied to the first layer, scaling the model by adding more layers would be no different than scaling a model built entirely with Dot-Product Attention. As such, the deeper the model becomes, the less relevant the increased overhead due to Neural Attention will be. 

In Table~\ref{tab:nlp_efficiency_comparison}, the memory requirements decrease sharply as the down-projection dimension is reduced. This is advantageous when considering the results in Table~\ref{tab:percent_improvement}, which show down-projection has little to no effect on perplexity improvements. Tables~\ref{tab:cifar10_efficiency_comparison} and~\ref{tab:cifar100_efficiency_comparison} show similar results during image classification training to those seen during NLP training.

\begin{table}[h!]
    \centering
    \caption{Memory usage and inference speed during training on Wikitext-103}
    \label{tab:nlp_efficiency_comparison}
    \resizebox{\textwidth}{!}{
        \begin{tabular}{|c|c|c|c|c|}
            \hline
            \textbf{Attention Type} & \textbf{Down-Projection $(d')$} & \textbf{Memory Required (GB/batch)} & \textbf{Inference Speed (ms/sample)} \\ \hline
            Dot-Product Attention & None ($d=64$) & 1.0 & 5.7 \\ \hline
            Neural Attention & 2 & 1.4 & 7.8 \\ \hline
            Neural Attention & 16 & 4.9 & 15.8 \\ \hline
            Neural Attention & None ($d=64$) & 8.7 & 48.1 \\ \hline
        \end{tabular}
    }
\end{table}

\begin{table}[h!]
    \centering
    \caption{Memory usage and inference speed during training on CIFAR-10}
    \label{tab:cifar10_efficiency_comparison}
    \resizebox{\textwidth}{!}{
        \begin{tabular}{|c|c|c|c|c|}
            \hline
            \textbf{Attention Type} & \textbf{Down-Projection $(d')$} & \textbf{Memory Required (GB/batch)} & \textbf{Inference Speed (ms/sample)} \\ \hline
            Dot-Product Attention & None ($d=64$) & 0.7 & 5.4 \\ \hline
            Neural Attention & 2 & 1.1 & 6.6 \\ \hline
        \end{tabular}
    }
\end{table}

\begin{table}[h!]
    \centering
    \caption{Memory usage and inference speed during training on CIFAR-100}
    \label{tab:cifar100_efficiency_comparison}
    \resizebox{\textwidth}{!}{
        \begin{tabular}{|c|c|c|c|c|}
            \hline
            \textbf{Attention Type} & \textbf{Down-Projection $(d')$} & \textbf{Memory Required (GB/batch)} & \textbf{Inference Speed (ms/sample)} \\ \hline
            Dot-Product Attention & None ($d=64$) & 0.7 & 5.4 \\ \hline
            Neural Attention & 2 & 1.1 & 6.6 \\ \hline
        \end{tabular}
    }
\end{table}

\section{Conclusion}
\label{sec:conclusion}
In this work, we have presented Neural Attention, a novel, modular, and easy-to-implement technique that has been shown to enhance the predictive capabilities of transformer models. By replacing dot products with feed-forward networks, Neural Attention captures local, nonlinear relationships, creating more expressive representations than standard methods can produce. We have demonstrated this enhanced expressivity through extensive experimentation in both NLP and vision contexts. Text generation tests on Wikitext-103 showed a reduction in perplexity of up to 2.19\%, while vision tests on CIFAR-10 and CIFAR-100 led to relative accuracy improvements of 3.0\% and 4.26\%, respectively. Notably, the additional overhead added by this technique was mitigated by our implementation, proving it to be both practical and scalable.

In future work, several research directions could further expand the impact of Neural Attention. When applied during the pretraining phase of large-scale conversational AI models, Neural Attention’s ability to enhance contextual understanding could lead to more coherent and nuanced responses in chat-based applications. Neural Attention may also find applications in edge computing, where its ability to increase performance with minimal loss of efficiency would be very valuable. Lastly, there is potential to further explore hybrid attention architectures, where Neural Attention is selectively applied in critical layers, exploring the optimal trade-off between computational efficiency and representational power. Given its modular design, strong empirical performance, and ease of integration into existing architectures, Neural Attention offers a promising new direction for advancing transformer-based AI models across diverse applications.

\bibliographystyle{abbrvnat}  

\bibliography{references}
\appendix

\section*{Appendix}

\section{Illustrating Dot-Product vs. Neural Attention}
\label{sec:Illustrations}
\begin{figure}[H]
    \centering
    \includegraphics[scale=0.48]{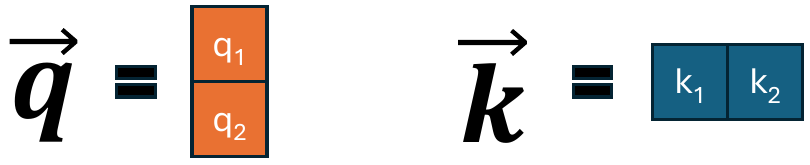}
    \caption{Embedding vectors $\vec{q}$ (shown in orange) and $\vec{k}$ (shown in  blue).}
    \label{fig:qandk}
\end{figure}

\begin{figure}[H]
    \centering
    \includegraphics[scale=0.52]{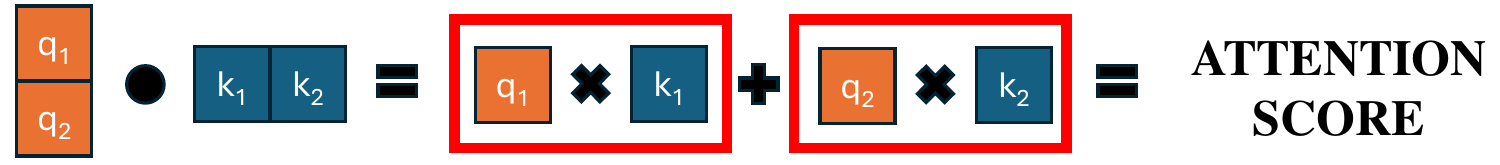}
    \caption{Dot product calculation between embedding vectors $\vec{q}$ (shown in orange) and $\vec{k}$ (shown in  blue).}
    \label{fig:dotproduct}
\end{figure}

\begin{figure}[H]
    \centering
    \includegraphics[scale=0.43]{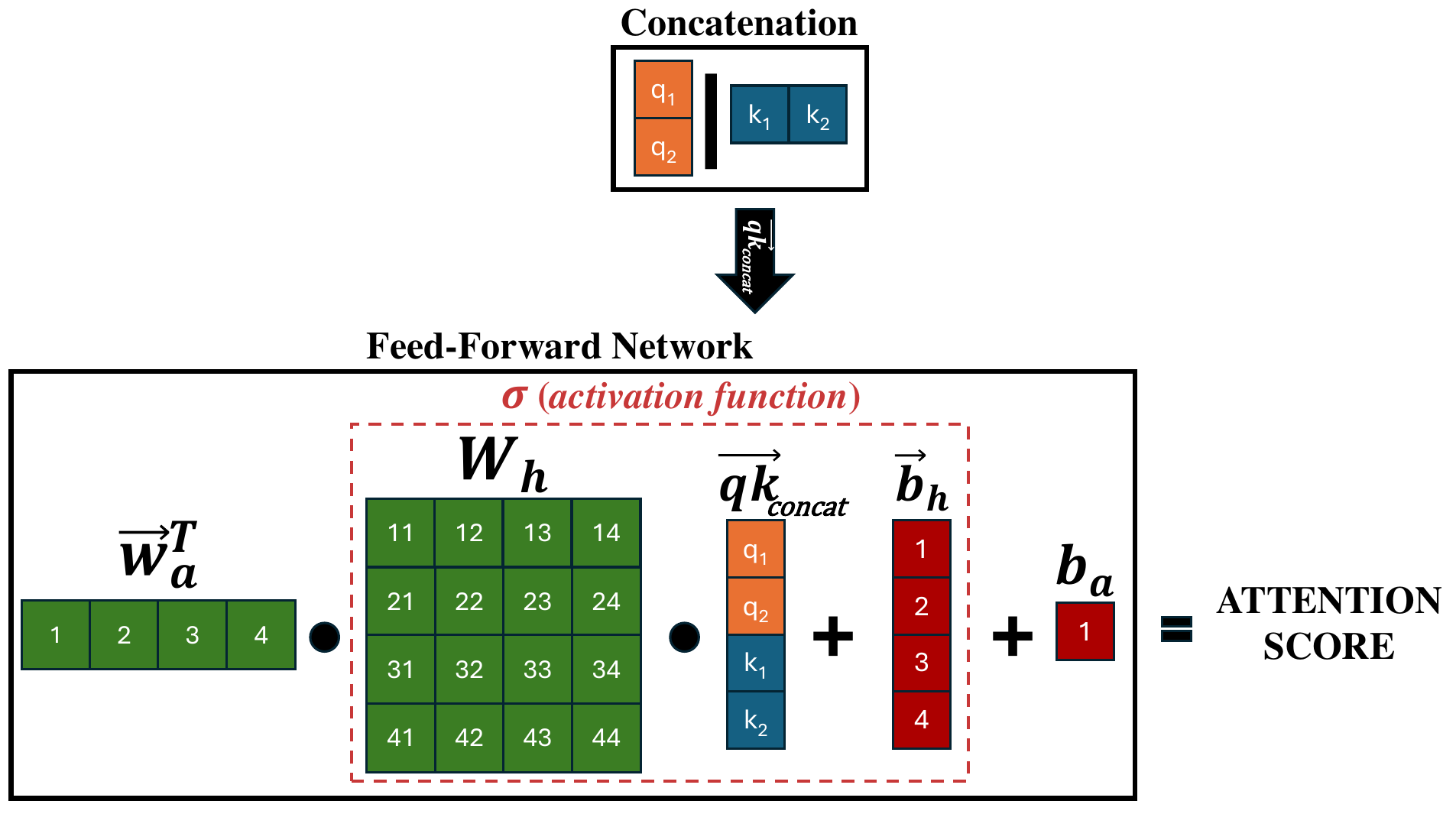}
    \caption{Neural Attention calculation between the two embedding vectors $\vec{q}$ (shown in orange) and $\vec{k}$ (shown in blue). Weights are shown in green, bias terms are shown in red, and a dashed box is used to represent the activation function.}
    \label{fig:neural_attention_score}
\end{figure}

Figure \ref{fig:qandk} shows two embedding vectors $\vec{q}$ and $\vec{k}$, each with an embedding dimension of $d=2$. We draw $\vec{q}$ as a column vector and $\vec{k}$ as a row vector to maintain familiar conventions when working with transformers.
In Figure~\ref{fig:dotproduct}, we demonstrate $\vec{q}$ and $\vec{k}$ producing an attention score through a dot product operation. Red boxes highlight how values in the embedding dimension interact.
In Figure~\ref{fig:neural_attention_score}, we demonstrate $\vec{q}$ and $\vec{k}$ producing an attention score through Neural Attention. The vectors are first concatenated before being transformed by a learnable weight matrix $\mathit{W_h}$, processed by an activation function, and then further transformed to a scalar value by a learnable weight vector $\vec{w}_a^\top$.
\begin{align*}
    \text{Attention Score} &= \vec{w}_a^\top \sigma\left(\mathit{W_h} \cdot \text{concat}(\vec{q}, \vec{k}) + \vec{b}_h\right) + b_a 
\end{align*}

\section{Example Neural Attention Calculation}
\label{sec:Example}
To give a specific example, consider applying Equation~\ref{eq:Attention_Matrix} to vector $\mathbf{C}_{11}$ from the tensor in Figure~\ref{fig:c_tensor}. The result can be seen below, along with a visual depiction in Figure~\ref{fig:attention_score_11}.

\begin{align*}
    \text{AttentionScore}_{11} = \vec{w}_a^\top \sigma(\mathit{W}_h \mathbf{C}_{ij} + \vec{b}_h) + b_a
\end{align*}

\begin{figure}[h!]
    \centering
      \includegraphics[scale=0.49]{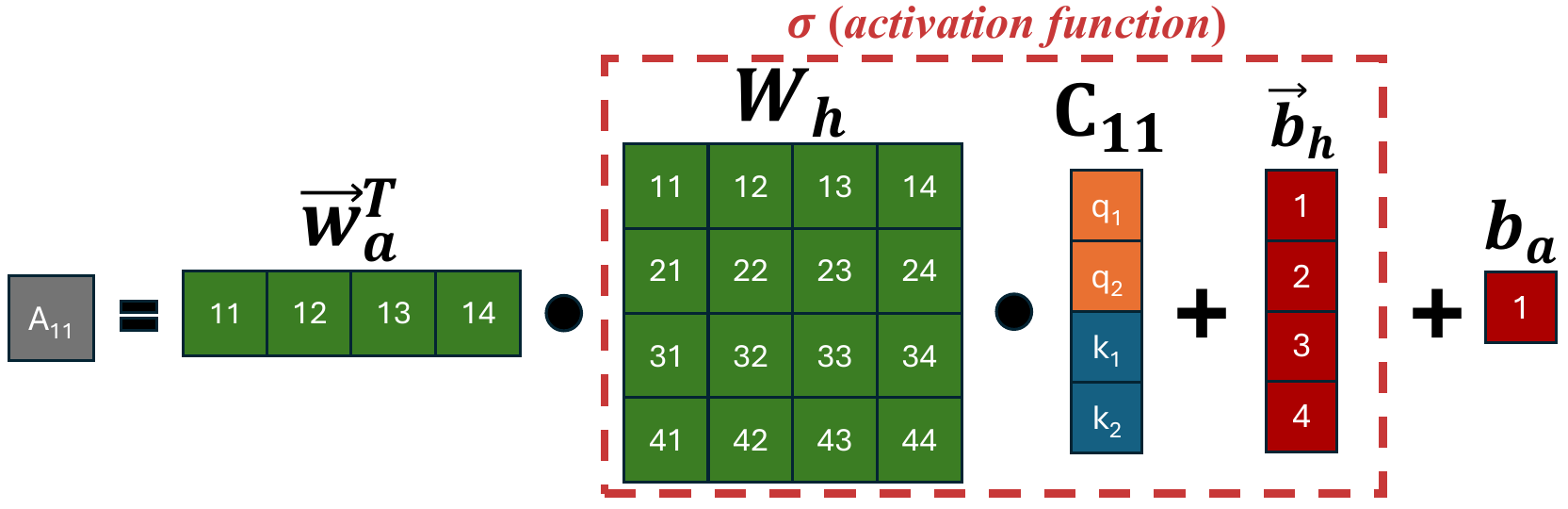}
    \caption{Visual representation of the calculation for $\text{AttentionScore}_{11}$, applying Equation~\ref{eq:Attention_Matrix} to $C_{11}$ of the tensor in Figure~\ref{fig:c_tensor}.}
    \label{fig:attention_score_11}
\end{figure}

\begin{figure}[H]
    \centering
      \includegraphics[scale=0.59]{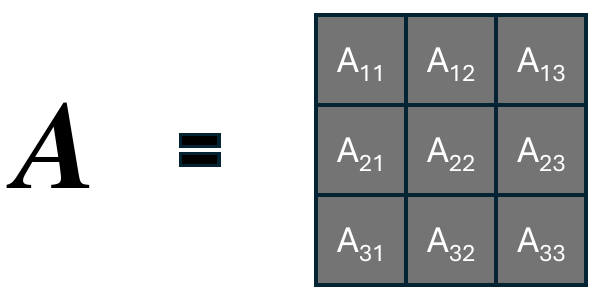}
    \caption{Matrix $\mathit{A}$ created by applying Equation~\ref{eq:Attention_Matrix} to the tensor $\mathbf{C}$ shown in Figure~\ref{fig:c_tensor}.}
    \label{fig:attention_matrix}
\end{figure}

If we extend what is depicted in Figure~\ref{fig:attention_score_11} to every $(i,j)$-th vector in tensor $\mathbf{C}$ shown in Figure~\ref{fig:c_tensor}, we can create the attention matrix $\mathit{A}$, shown in Figure~\ref{fig:attention_matrix}.

\section{Training Graphs}
\label{sec:graphs}
\begin{figure}[H]
    \centering
      \includegraphics[scale=.7]{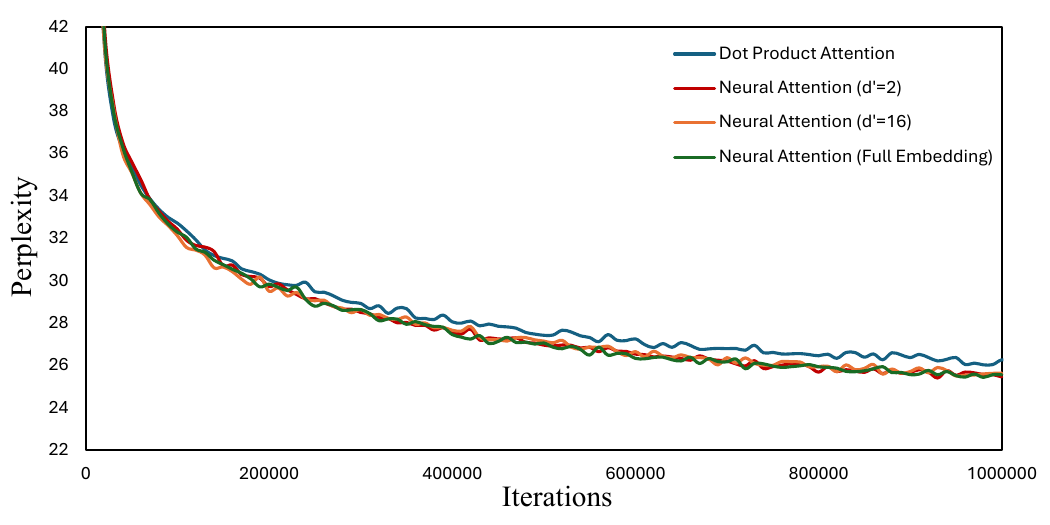}
    \caption{Perplexity comparison of Dot-Product Attention and Neural Attention across training iterations on WikiText-103.}
    \label{fig:NLP_Graph}
\end{figure}

\begin{figure}[H]
    \centering
      \includegraphics[scale=.7]{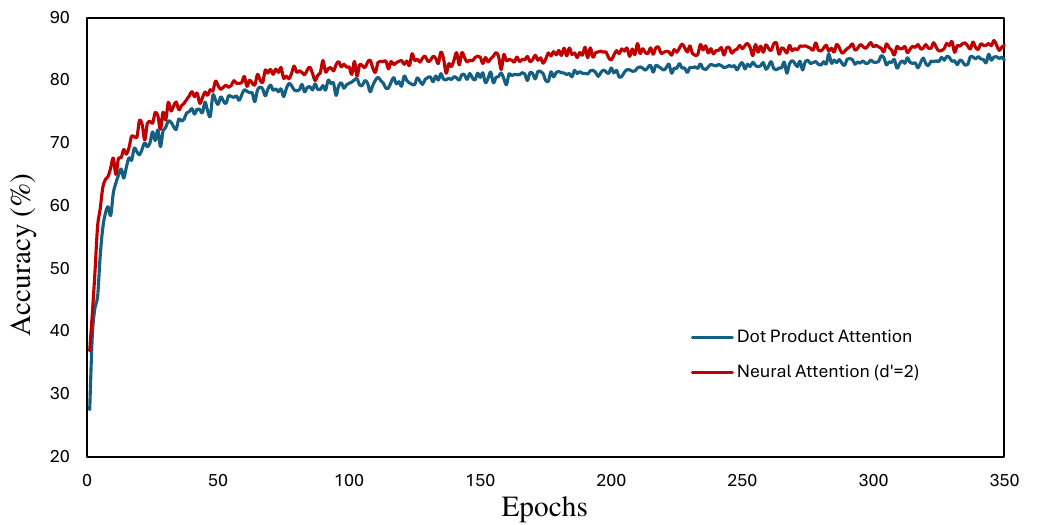}
    \caption{Accuracy comparison of Dot-Product Attention and Neural Attention across training epochs on CIFAR-10.}
    \label{fig:CIFAR10_Graph}
\end{figure}

\begin{figure}[H]
    \centering
      \includegraphics[scale=.7]{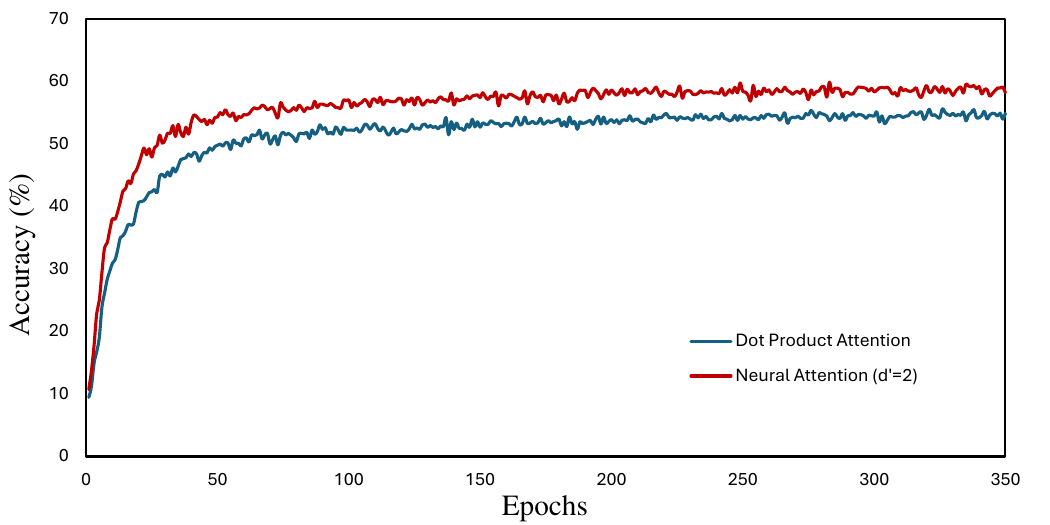}
    \caption{Accuracy comparison of Dot-Product Attention and Neural Attention across training epochs on CIFAR-100.}
    \label{fig:CIFAR100_Graph}
\end{figure}

\end{document}